\documentclass[10pt,twocolumn,letterpaper]{article}

\usepackage{cvpr}
\usepackage{times}
\usepackage{epsfig}
\usepackage{graphicx}
\usepackage{amsmath}
\usepackage{amssymb}
\usepackage{bm}
\usepackage{multirow}
\usepackage{nicefrac}       
\usepackage{fixltx2e}
\usepackage{array}
\usepackage{algorithm, algpseudocode}
\usepackage{mathtools}
\usepackage{mathrsfs}
\usepackage{amsthm}

\usepackage{etoolbox}
\makeatletter
\patchcmd\@combinedblfloats{\box\@outputbox}{\unvbox\@outputbox}{}{%
   \errmessage{\noexpand\@combinedblfloats could not be patched}%
}%
 \makeatother

\newtheorem{theorem}{Theorem}

\let\oldReturn\Return
\renewcommand{\Return}{\State\oldReturn}

\usepackage[pagebackref=true,breaklinks=true,letterpaper=true,colorlinks,bookmarks=false]{hyperref}

\cvprfinalcopy 

\def\cvprPaperID{2676} 

\begin{document}

\title{Efficient Decision-based Black-box Adversarial Attacks on Face Recognition}

\author{Yinpeng Dong$^{1}$, Hang Su$^{1}$, Baoyuan Wu$^{2}$, Zhifeng Li$^{2}$, Wei Liu$^{2}$, Tong Zhang$^{3}$, Jun Zhu$^{1}$\thanks{Corresponding author.}\\
$^{1}$ Dept. of Comp. Sci. and Tech., BNRist Center, State Key Lab for Intell. Tech. \& Sys.,\\
$^{1}$ Institute for AI, THBI Lab, Tsinghua University, Beijing, 100084, China\\
$^{2}$ Tencent AI Lab \hspace{3ex} $^{3}$ Hong Kong University of Science and Technology \\
\small{dyp17@mails.tsinghua.edu.cn, suhangss@mail.tsinghua.edu.cn, wubaoyuan1987@gmail.com} \\
\small{michaelzfli@tencent.com, wl2223@columbia.edu, tongzhang@tongzhang-ml.org, dcszj@mail.tsinghua.edu.cn}}

\maketitle

\begin{abstract}
   Face recognition has obtained remarkable progress in recent years due to the great improvement of deep convolutional neural networks (CNNs). However, deep CNNs are vulnerable to adversarial examples, which can cause fateful consequences in real-world face recognition applications with security-sensitive purposes. Adversarial attacks are widely studied as they can identify the vulnerability of the models before they are deployed. In this paper, we evaluate the robustness of state-of-the-art face recognition models in the decision-based black-box attack setting, where the attackers have no access to the model parameters and gradients, but can only acquire hard-label predictions by sending queries to the target model. This attack setting is more practical in real-world face recognition systems. To improve the efficiency of previous methods, we propose an evolutionary attack algorithm, which can model the local geometries of the search directions and reduce the dimension of the search space. Extensive experiments demonstrate the effectiveness of the proposed method that induces a minimum perturbation to an input face image with fewer queries. We also apply the proposed method to attack a real-world face recognition system successfully.
\end{abstract}

\section{Introduction}

Recent progress in deep convolutional neural networks (CNNs)~\cite{simonyan2014very,szegedy2015going,he2015deep} has led to substantial performance improvements in a broad range of computer vision tasks. 
Face recognition, as one of the most important computer vision tasks, has been greatly facilitated by deep CNNs~\cite{schroff2015facenet,wen2016discriminative,liu2017sphereface,wang2018cosface,deng2018arcface,wang2018orthogonal,cvpr19}.
There are usually two sub-tasks in face recognition: face verification and face identification~\cite{huang2008labeled,kemelmacher2016megaface}. The former distinguishes whether a pair of face images represent the same identity, while the latter classifies an image to an identity. The state-of-the-art face recognition models realize these two tasks by using deep CNNs to extract face features that have minimum intra-class variance and maximum inter-class variance.
Due to the excellent performance of these models, face recognition has been widely used for identity authentication in enormous applications, such as finance/payment, public access, criminal identification, \etc.

Despite the great success in various applications, deep CNNs are known to be vulnerable to adversarial examples~\cite{szegedy2013intriguing,goodfellow2014explaining,Moosavi2016Universal,Dong_2018_CVPR}. These maliciously generated adversarial examples are often indistinguishable from legitimate ones for human observers by adding small perturbations. 
But they can make deep models produce incorrect predictions.
The face recognition systems based on deep CNNs have also been shown their vulnerability against such adversarial examples.
For instance, adversarial perturbations can be made to the eyeglass that, when worn, allows attackers to evade being recognized or impersonate another individual~\cite{Sharif2016Accessorize,Sharif2017Adversarial}.
The insecurity of face recognition systems in real-world applications, especially those with sensitive purposes, can cause severe consequences and security issues.

\begin{figure}[t]
  \centering
    \includegraphics[width=1.0\linewidth]{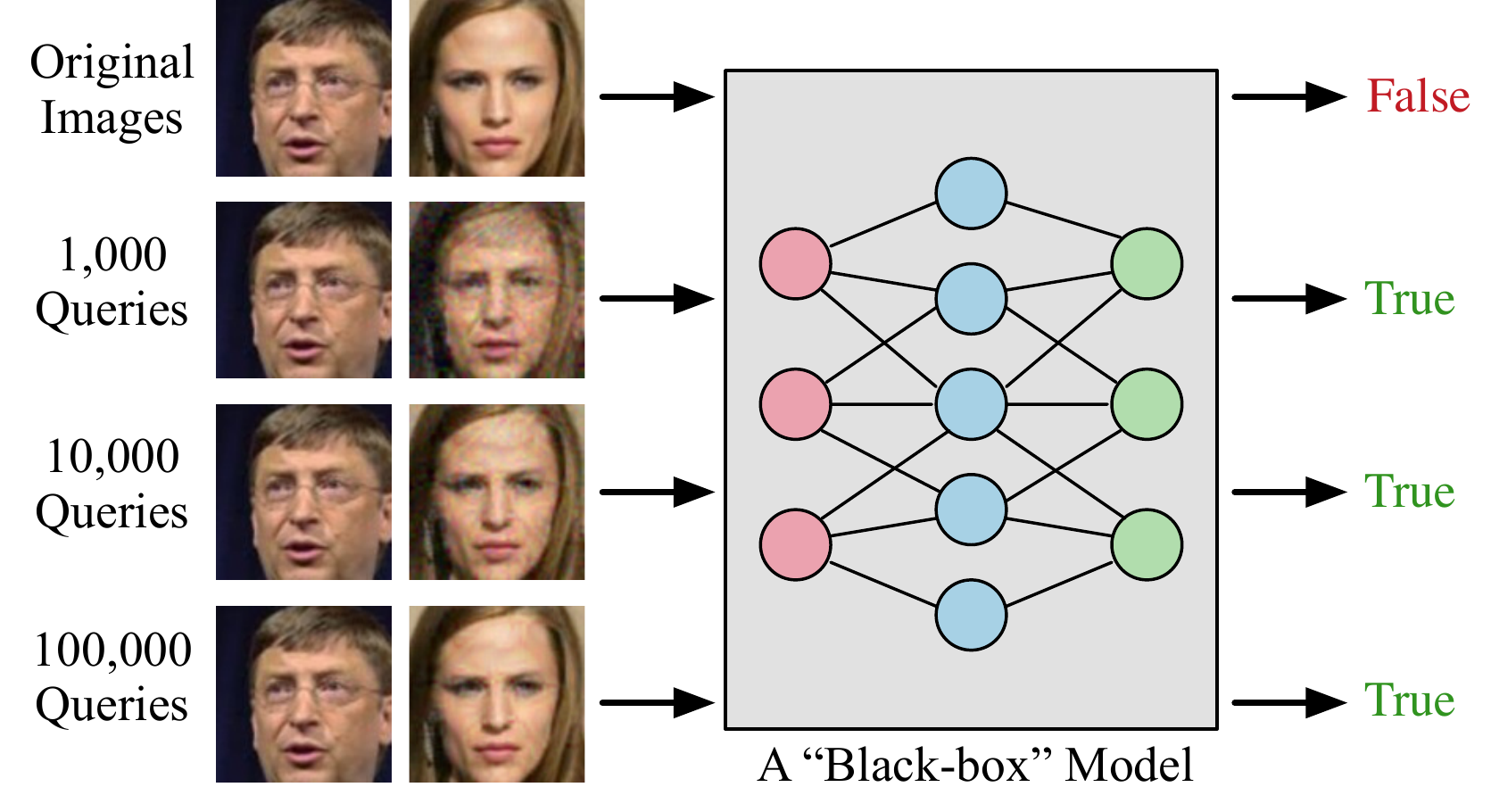}
    \caption{Demonstration of the decision-based black-box attack setting. Given a black-box model, the attackers use queries to generate adversarial examples with minimum perturbations.}
    \label{fig:demo}
    \vspace{-2.7ex}
\end{figure}

To evaluate the robustness of face recognition systems in real-world applications, adversarial attacks can serve as an important surrogate, as they can identify the vulnerability of these systems~\cite{carlini2017towards} and help improve the robustness~\cite{goodfellow2014explaining,Madry2017Towards}.
However, existing attack methods~\cite{Sharif2016Accessorize,Sharif2017Adversarial} for face recognition are mainly based on the \textit{white-box} scenario, where the attackers know the internal structure and parameters of the system being attacked.
Accordingly, the attack objective function can be directly optimized by gradient-based methods.
This setting is clearly impractical in real-world cases, when the attackers cannot get access to the model details.
Instead, we focus on a more realistic and general \textit{decision-based black-box} setting~\cite{Brendel2018Decision}, where no model information is exposed except that the attackers can only query the target model and obtain corresponding hard-label predictions.
The goal of attacks is to generate adversarial examples with minimum perturbations by limited queries. 
This attack scenario is much more challenging, since the gradient cannot be directly computed and the predicted probability is not provided.
On the other hand, it is much more realistic and important, because most of the real-world face recognition systems are black-box and only provide hard-label outputs. To the best of our knowledge, it is the first attempt to conduct adversarial attacks on face recognition in this setting.

Several methods~\cite{Brendel2018Decision,ilyas2018black,cheng2018query} have been proposed to perform decision-based black-box attacks. However, they lack the efficiency in the sense that they usually require a tremendous number of queries to converge, or get a relatively large perturbation given a limited budget of queries.
Therefore, we consider how to efficiently generate adversarial examples for decision-based black-box attacks by inducing a smaller perturbation to each sample with fewer queries. 

To address the aforementioned issues, we propose an \textbf{evolutionary attack} method for query-efficient adversarial attacks in the decision-based black-box setting. 
Given the attack objective function, the proposed method can optimize it in the black-box manner through queries solely.
Our method can find better search directions by modeling their local geometries. It further improves the efficiency by reducing the dimension of the search space. 
We apply the proposed method to comprehensively study the robustness of several state-of-the-art face recognition models, including SphereFace~\cite{liu2017sphereface}, CosFace~\cite{wang2018cosface}, and ArcFace~\cite{deng2018arcface}, under the decision-based black-box scenario.
Extensive experiments conducted on the most popular public-domain face recognition datasets such as Labeled Face in the Wild (LFW)~\cite{huang2008labeled} and MegaFace Challenge~\cite{kemelmacher2016megaface} demonstrate the effectiveness of the proposed method.
We further apply our method to attack a real-world face recognition system to show its practical applicability.
In summary, our major contributions are:
\begin{itemize}
    \item We propose a novel evolutionary attack method under the decision-based black-box scenario, which can model the local geometries of the search directions and meanwhile reduce the dimension of the search space. The evolutionary attack method is generally applicable for any image recognition task, and significantly improves the efficiency over existing methods.\vspace{-1.5ex}
    \item We thoroughly evaluate the robustness of several state-of-the-art face recognition models by decision-based black-box attacks in various settings. We demonstrate the vulnerability of these face models in this setting.\vspace{-1.5ex}
    \item We show the practical applicability of the proposed method by successfully attacking a real-world face recognition system.
\end{itemize}

\section{Related Work}
\textbf{Deep face recognition.} 
DeepFace~\cite{taigman2014deepface} and DeepID~\cite{sun2014deep} treat face recognition as a multi-class classification problem and use deep CNNs to learn features supervised by the softmax loss.
Triplet loss~\cite{schroff2015facenet} and center loss~\cite{wen2016discriminative} are proposed to increase the Euclidean margin in the feature space between classes.
The angular softmax loss is proposed in SphereFace~\cite{liu2017sphereface} to learn angularly discriminative features.
CosFace~\cite{wang2018cosface} uses the large margin cosine loss to maximize the cosine margin.
The additive angular margin loss is proposed in ArcFace~\cite{deng2018arcface} to learn highly discriminative features.

\textbf{Adversarial attacks on face recognition.} Deep CNNs are highly vulnerable to adversarial examples~\cite{szegedy2013intriguing,goodfellow2014explaining,Moosavi2016Universal}. Face recognition has also been shown the vulnerability against attacks.
In~\cite{Sharif2016Accessorize}, the perturbations are constrained to the eyeglass region and generated by gradient-based methods, which fool face recognition systems even in the physical world.
The adversarial eyeglasses can also be produced by generative networks~\cite{Sharif2017Adversarial}.
However, these methods rely on the white-box manipulations of face recognition models, which is unrealistic in real-world applications.
Instead, we focus on evaluating the robustness of face recognition models in the decision-based black-box attack setting.

\textbf{Black-box attacks.} Black-box attacks can be divided into transfer-based, score-based, and decision-based attacks.
Transfer-based attacks generate adversarial examples for a white-box model and attack the black-box model based on the transferability~\cite{liu2016delving,Dong_2018_CVPR}.
In score-based attacks, the predicted probability is given by the model. Several methods rely on approximated gradients to generate adversarial examples~\cite{chen2017zoo,ilyas2018black}.
In decision-based attacks, we can only obtain the hard-label predictions. 
The boundary attack method is based on random walks on the decision boundary~\cite{Brendel2018Decision}.
The optimization-based method~\cite{cheng2018query} formulates this problem as a continuous optimization problem and estimates the gradient for optimization. However, it needs to calculate the distance to the decision boundary along a direction by binary search.
In~\cite{ilyas2018black}, the predicted probability is estimated by hard-label predictions. Then, the natural evolution strategy (NES) is used to maximize the target class probability or minimize the true class probability.
These methods generally require a large number of queries to generate an adversarial example with a minimum perturbation, or converge to a large perturbation with few queries .

\section{Methodology}
In this section, we first introduce the decision-based black-box attack setting against a face recognition model, and then detail the proposed evolutionary attack method.

\subsection{Attack Setting}
Let $f(\bm{x}):\mathcal{X} \rightarrow \mathcal{Y}$ ($\mathcal{X}\subset\mathbb{R}^n$) denote the face recognition model that predicts a label for an input face image. 
For face verification, the model relies on another face image to identify whether the pair of images belong to the same identity, and outputs a binary label in $\mathcal{Y}=\{0,1\}$. 
For face identification, the model $f(\bm{x})$ compares the input image $\bm{x}$ with a gallery set of face images, and then classifies $\bm{x}$ as a specific identity. So it can be viewed as a multi-class classification task, where $\mathcal{Y} = \{1,2,...,K\}$, with $K$ being the number of identities.
Although the face recognition model $f(\bm{x})$ uses an additional face image or a set of face images for recognizing $\bm{x}$, we do not explicitly describe the dependency of $f(\bm{x})$ on the compared images for simplicity.

Given a real face image $\bm{x}$, the goal of attacks is to generate an adversarial face image $\bm{x}^*$ in the vicinity of $\bm{x}$ but is misclassified by the model.
It can be obtained by solving a constrained optimization problem
\begin{equation}
\min_{\bm{x}^*} \mathcal{D}(\bm{x}^*, \bm{x}), \hspace{0.2cm} \text{s.t.} \hspace{0.1cm}\mathcal{C}(f(\bm{x}^*))=1,
\label{eq:1}
\end{equation}
where $\mathcal{D}(\cdot,\cdot)$ is a distance metric, and $\mathcal{C}(\cdot)$ is an adversarial criterion that takes $1$ if the attack requirement is satisfied and $0$ otherwise.
We use the $L_2$ distance as $\mathcal{D}$. 
The constrained problem in Eq.~\eqref{eq:1} can be equivalently reformulated as the following unconstrained optimization problem 
\begin{equation}
\min_{\bm{x}^*} \mathcal{L}(\bm{x}^*) = \mathcal{D}(\bm{x}^*, \bm{x}) + \delta\big(\mathcal{C}(f(\bm{x}^*)) = 1\big),
\label{eq:2}
\end{equation}
where $\delta(a) = 0$ if $a$ is true, otherwise $\delta(a) = +\infty$. 
By optimizing Eq.~\eqref{eq:2}, we can obtain an image $\bm{x}^*$ with a minimum perturbation, which is also adversarial according to the criterion.
Note that in the above objective functions, $\mathcal{C}$ cannot be defined as a continuous criterion such as cross-entropy loss, since the model $f(\bm{x})$ only gives discrete hard-label outputs in this problem.
In particular, we specify $\mathcal{C}$ according to the following two types of attacks.

\textbf{Dodging} attack corresponds to generating an adversarial image that is recognized wrong or not recognized. Dodging attack could be used to protect personal privacy against excessive surveillance. 
For face verification, given a pair of face images belonging to the same identity, the attacker seeks to modify one image and makes the model recognize them as not the same identity. So the criterion is $\mathcal{C}(f(\bm{x}^*)) = \mathbb{I}(f(\bm{x}^*) = 0)$, where $\mathbb{I}$ is the indicator function.
For face identification, the attacker generates an adversarial face image with the purpose that it is recognized as any other identity. The criterion is $\mathcal{C}(f(\bm{x}^*)) = \mathbb{I}(f(\bm{x}^*) \neq y)$, where $y$ is the true identity of the real image $\bm{x}$.

\textbf{Impersonation} attack works as seeking an adversarial image recognized as a specific identity, which could be used to evade the face authentication systems.
For face verification, the attacker tries to find an adversarial image that is recognized as the same identity of another image, while the original images are not from the same identity. The criterion is $\mathcal{C}(f(\bm{x}^*)) = \mathbb{I}(f(\bm{x}^*) = 1)$.
For face identification, the generated adversarial image needs to be classified as a specific identity $y^*$, so $\mathcal{C}(f(\bm{x}^*)) = \mathbb{I}(f(\bm{x}^*) = y^*)$.

\subsection{Evolutionary Attack}

Since we cannot get access to the configuration and parameters of $f(\bm{x})$ but can only send queries to probe the model, we resort to black-box optimization techniques to minimize the objective function in Eq.~\eqref{eq:2}. 
Gradient estimation methods~\cite{Nesterov2017Random,Ghadimi2012Stochastic,Flaxman2005Online} approximate the gradient of the objective function by finite difference and update the solution by gradient descent, which are commonly used for score-based black-box attacks, when the predicted probability is given by the model~\cite{chen2017zoo,ilyas2018black}.
However, in the case of hard-label output, the attack objective function is discontinuous and the output is insensitive to small input perturbations.
So the gradient estimation methods cannot be directly used.
Some methods~\cite{cheng2018query,ilyas2018black} successfully reformulate the discontinuous optimization problem in Eq.~\eqref{eq:2} as some continuous optimization problems and use gradient estimation methods for optimization. But they need to calculate the distance of a point to the decision boundary or estimate the predicted probability by the hard-label outputs, which are less efficient as demonstrated in the experiments.
Therefore, we consider how to directly optimize Eq.~\eqref{eq:2} efficiently.

In this paper, we propose a novel \textbf{evolutionary attack} method to solve the black-box optimization problem. 
Our method is based on a simple and efficient variant of covariance matrix adaptation evolution strategy (CMA-ES)~\cite{hansen2001completely}, which is the (1+1)-CMA-ES~\cite{igel2006computational}.
In each update iteration of the (1+1)-CMA-ES, a new offspring (candidate solution) is generated from its parent (current solution) by adding a random noise, the objective of these two solutions are evaluated, and the better one is selected for the next iteration.
This method is capable for solving black-box optimization problems.
However, directly applying the (1+1)-CMA-ES to optimize Eq.~\eqref{eq:2} is inefficient due to the high dimension of $\bm{x}^*$. 
Considering the query limit in decision-based black-box attacks for face images, the original (1+1)-CMA-ES may be infeasible. 
To accelerate this algorithm, we design an appropriate distribution to sample the random noise in each iteration, which can model the local geometries of the search directions. We also propose several techniques to reduce the dimension of the search space by considering the special characteristics of this problem.

\begin{algorithm}[!t]
\small
\caption{The evolutionary attack algorithm}
\label{alg:evolution}
\begin{algorithmic}[1]
\Require The attack objective function $\mathcal{L}(\bm{x}^*)$; the original face image $\bm{x}$; the dimension $n\in\mathbb{N}_{+}$ of the input space ($\bm{x}^* \in \mathbb{R}^n$); the dimension $m\in\mathbb{N}_{+}$ of the search space; the number of coordinates $k\in\mathbb{N}_{+}$ for stochastic coordinate selection.
\Require The total number of queries $T$.
\State Initialize $\mathbf{C} = \mathbf{I}_{m}$, $\bm{p}_c = \mathbf{0}$, $\sigma,\mu,c_c,c_{cov} \in \mathbb{R}_{+}$, $\tilde{\bm{x}}^* \in \mathbb{R}^n$;
\For {$t = 1$ to $T$}
\State Sample $\bm{z} \sim \mathcal{N}(\mathbf{0}, \sigma^2\mathbf{C})$;
\State Select $k$ coordinates from $m$ with probability proportional to each diagonal element in $\mathbf{C}$;
\State Set the non-selected coordinates of $\bm{z}$ to $0$;
\State Upscale $\bm{z}$ to $\mathbb{R}^n$ by bilinear interpolation and obtain $\tilde{\bm{z}}$;
\State $\tilde{\bm{z}} \leftarrow \tilde{\bm{z}} + \mu(\bm{x}-\tilde{\bm{x}}^*)$;
\If {$\mathcal{L}(\tilde{\bm{x}}^* + \tilde{\bm{z}}) < \mathcal{L}(\tilde{\bm{x}}^*)$}
\State $\tilde{\bm{x}}^* \leftarrow \tilde{\bm{x}}^* + \tilde{\bm{z}}$;
\State Update $\bm{p}_c$ and $\mathbf{C}$ by $\bm{z}$ according to Eq.~\eqref{eq:ep} and Eq.~\eqref{eq:cma};
\EndIf
\EndFor
\Return $\tilde{\bm{x}}^*$.
\end{algorithmic}
\end{algorithm}

The overall evolutionary attack algorithm is outlined in Algorithm~\ref{alg:evolution}.
Rather than the original $n$-dimensional input space, we perform search in a lower dimensional space $\mathbb{R}^{m}$ with $m < n$.
In each iteration, we first sample a random vector $\bm{z}$ from $\mathcal{N}(\mathbf{0}, \sigma^2\mathbf{C})$ such that $\bm{z}\in\mathbb{R}^{m}$, where $\mathbf{C}$ is a diagonal covariance matrix to model the local geometries of the search directions. 
We then select $k$ coordinates randomly for search, according to the assumption that only a fraction of pixels are important for finding an adversarial image. We keep the value of the selected $k$ coordinates of $\bm{z}$ by setting the others to $0$. 
We upscale $\bm{z}$ to the input space by bilinear interpolation and get $\tilde{\bm{z}} \in \mathbb{R}^{n}$.
We further add a bias to $\tilde{\bm{z}}$ to minimize the distance between the adversarial and original images. We finally test whether we get a better solution. If we indeed find a better solution, we jump to it and update the covariance matrix.
In the following, we will give a detailed description of each step in the algorithm.

\vspace{-1.8ex}
\subsubsection{Initialization}

In Algorithm~\ref{alg:evolution}, $\tilde{\bm{x}}^*$ should be initialized at first (in Step 1).
If the initial $\tilde{\bm{x}}^*$ does not satisfy the adversarial criterion, $\mathcal{L}(\tilde{\bm{x}}^*)$ equals to $+\infty$.
For subsequent iterations, adding a random vector can rarely make the search point adversarial due to that deep CNNs are generally robust to random noises~\cite{szegedy2013intriguing}, and thus the loss function will keep being $+\infty$. So we initialize $\tilde{\bm{x}}^*$ with a sample that already satisfies the adversarial criterion.
The following updates will also keep $\tilde{\bm{x}}^*$ adversarial, and at the same time minimize the distance between $\tilde{\bm{x}}^*$ and $\bm{x}$.
For dodging attack, the initial $\tilde{\bm{x}}^*$ can be simply set as a random vector.
For impersonation attack, we use the target image as the initial point of $\tilde{\bm{x}}^*$.

\vspace{-1.8ex}
\subsubsection{Mean of Gaussian Distribution}
\label{sec:bias}
We explain why we need to add a bias term to the random vector in Step 7.
Assume now that the dimension of the search space is the same as that of the input space and we select all coordinates for search (\ie, $k=m=n$).
In each iteration, a random vector $\bm{z}$ is sampled from a Gaussian distribution. 
In general, the distribution should be unbiased (with zero mean) for better exploration in the search space.
But in our problem, sampling the random vector from a zero mean Gaussian distribution will result in nearly zero probability of updates as $n\rightarrow\infty$, given by Theorem~\ref{Theorem:0}.
\vspace{-1ex}
\begin{theorem}\label{Theorem:0}
(Proof in Appendix A) Assume that the covariance matrix $\mathbf{C}$ is positive definite. Let $\lambda_{max}$ and $\lambda_{min} (>0)$ be the largest and smallest eigenvalues of $\mathbf{C}$, respectively. Then, we have
\vspace{-1ex}
\begin{equation*}
P_{\bm{z}\sim\mathcal{N}(\mathbf{0}, \sigma^2\mathbf{C})}\big( \mathcal{L}(\tilde{\bm{x}}^* + \bm{z}) < \mathcal{L}(\tilde{\bm{x}}^*)\big) \leq \frac{4\lambda_{max}\|\tilde{\bm{x}}^*-\bm{x}\|^2}{\sigma^2\lambda_{min}^2n^2}.
\end{equation*}
\end{theorem}
\vspace{-1ex}
From Theorem~\ref{Theorem:0}, we need to draw $\mathcal{O}(n^2)$ samples from the zero mean Gaussian distribution for only one successful update, which is inefficient and costly when $n$ is large.
This happens because in high dimensional search space, a randomly drawn vector $\bm{z}$ is almost orthogonal to $\tilde{\bm{x}}^*-\bm{x}$, thus the distance $\mathcal{D}(\tilde{\bm{x}}^* + \bm{z}, \bm{x})$ will be rarely smaller than $\mathcal{D}(\tilde{\bm{x}}^*, \bm{x})$.
To address this problem, the random vector $\bm{z}$ should be sampled from a biased distribution towards minimizing the distance of $\tilde{\bm{x}}^*$ from the original image $\bm{x}$.
So we add a bias term $\mu(\bm{x} - \tilde{\bm{x}}^*)$ to $\tilde{\bm{z}}$ (the same as $\bm{z}$ when $k=m=n$) in Step 7, where $\mu$ is a critical hyper-parameter controlling the strength of going towards the original image $\bm{x}$. We will specify the update procedure of $\mu$ in Sec.~\ref{sec:alg}.

\begin{table*}[!t]
\footnotesize
\begin{center}
\begin{tabular}{c|c|cccc|cccc|cccc}

\hline
\multicolumn{2}{c|}{Model} & \multicolumn{4}{c|}{SphereFace~\cite{liu2017sphereface}} &  \multicolumn{4}{c|}{CosFace~\cite{wang2018cosface}} & \multicolumn{4}{c}{ArcFace~\cite{deng2018arcface}} \\
\hline
\multicolumn{2}{c|}{Queries} & 1,000 & 5,000 & 10,000 & 100,000 & 1,000 & 5,000 & 10,000 & 100,000 & 1,000 & 5,000 & 10,000 & 100,000 \\
\hline\hline
\multirow{4}{*}{Dodging} & Boundary~\cite{Brendel2018Decision} & 2.3e-2 & 9.3e-3 & 7.0e-4 & 1.9e-5 & 2.0e-2 & 7.5e-3 & 7.7e-4 & 1.6e-5 & 2.4e-2 & 1.6e-2 & 1.5e-3 & 2.3e-5\\
& Optimization~\cite{cheng2018query} & 1.2e-2 & 2.9e-3 & 1.3e-3 & 7.1e-5 & 1.1e-2 & 2.9e-3 & 1.3e-3 & 6.6e-5 & 1.5e-2 & 5.4e-3 & 2.6e-3 & 9.9e-5\\
& NES-LO~\cite{ilyas2018black} & 1.4e-1 & 3.8e-2 & 2.4e-2 & 7.4e-3 & 1.4e-1 & 3.5e-2 & 2.0e-2 & 6.5e-3 & 1.4e-1 & 3.9e-2 & 2.3e-2 & 1.5e-2\\
& Evolutionary & \textbf{1.6e-3} & \textbf{8.9e-5} & \textbf{3.4e-5} & \textbf{1.3e-5} & \textbf{1.7e-3} & \textbf{9.1e-5} & \textbf{3.3e-5} & \textbf{1.1e-5} & \textbf{2.8e-3} & \textbf{1.5e-4} & \textbf{5.2e-5} & \textbf{1.6e-5}\\
\hline
\multirow{4}{*}{Impersonation} & Boundary~\cite{Brendel2018Decision} & 1.5e-2 & 6.3e-3 & 5.7e-4 & 1.6e-5 & 1.1e-2 & 2.9e-3 & 2.8e-4 & 7.4e-6 & 2.0e-2 & 9.2e-3 & 1.2e-3 & 1.7e-5\\
& Optimization~\cite{cheng2018query} & 1.1e-2 & 3.3e-3 & 1.3e-3 & 6.1e-5 & 7.7e-3 & 1.9e-3 & 7.1e-4& 2.8e-5 & 1.6e-2 & 7.0e-3 & 3.3e-3 & 7.7e-5\\
& NES-LO~\cite{ilyas2018black} & 8.4e-2 & 2.6e-2 & 1.7e-2 & 5.5e-3 & 9.3e-2 & 2.0e-2 & 1.2e-2 & 3.1e-3 & 9.3e-2 & 3.0e-2 & 1.9e-2 & 8.1e-3\\
& Evolutionary & \textbf{1.2e-3} & \textbf{7.2e-5} & \textbf{2.9e-5} & \textbf{1.2e-5} & \textbf{6.5e-4} & \textbf{3.7e-5} & \textbf{1.5e-5} & \textbf{5.3e-6} & \textbf{2.3e-3} & \textbf{1.2e-4} & \textbf{3.9e-5} & \textbf{1.2e-5}\\
\hline
\end{tabular}
\end{center}
\vspace{-2ex}
\caption{The results on face verification conducted on the LFW dataset. We report the average distortions (MSEs) of the adversarial images generated by different methods for SphereFace, CosFace, and ArcFace given 1,000, 5,000, 10,000, and 100,000 queries.}
\label{tab:verification}
\vspace{-3.5ex}
\end{table*}

\vspace{-1.8ex}
\subsubsection{Covariance Matrix Adaptation}
\label{sec:CMA}
The adaptation of covariance matrix $\mathbf{C}$ is suitable for solving non-separable optimization problems since it can model the local geometries of the search directions~\cite{hansen2001completely}.
For example, an appropriately set covariance matrix can make the random vectors generated predominantly in the direction of narrow valleys.
In learning all pair-wise dependencies between dimensions, the storage and computation complexity of the covariance matrix is at least $\mathcal{O}(m^2)$, which is unacceptable when $m$ is large.
For black-box adversarial attacks, the dimension of the search space is extremely large (\eg, $m=45\times45\times3$ in our experiments).
Therefore, we relax the covariance matrix to be a diagonal matrix for efficient computation.
Inspired by~\cite{ros2008simple} which uses a diagonal covariance matrix for CMA-ES, 
we design an update rule for the adaptation of the diagonal covariance matrix $\mathbf{C}$ (in Step 10) after each successful trial as
\vspace{-1ex}
\begin{equation}
\label{eq:ep}
\bm{p}_c = (1-c_c)\bm{p}_c + \sqrt{c_c(2-c_c)} \frac{\bm{z}}{\sigma},
\end{equation}\vspace{-3ex}
\begin{equation}
\label{eq:cma}
c_{ii} = (1-c_{cov}) c_{ii} + c_{cov} (\bm{p}_c)_i^2,
\end{equation}
where $\bm{p}_c \in \mathbb{R}^m$ is called the evolution path as it stores the exponentially decayed successful search directions; for $i = 1, ..., m$, $c_{ii}$ is the diagonal element of $\mathbf{C}$ and ($\bm{p}_c)_i$ is the $i$-th element of $\bm{p}_c$. $c_c$ and $c_{cov}$ are two hyper-parameters of CMA.
An intuitive explanation of this update is that the variance along the past successful directions should be enlarged for future search.

\vspace{-1.8ex}
\subsubsection{Stochastic Coordinate Selection}
\label{sec:SCS}
For adversarial attacks, the perturbations added to the images could be very sparse to fool deep CNNs~\cite{su2017one}, indicating that only a fraction of coordinates (pixels) are sufficient for finding the adversarial images.
We can also accelerate the black-box optimization if we could identify the important coordinates.
However, this is non-trivial in the decision-based black-box attack setting.
Fortunately, our algorithm provides a natural way to find the useful coordinates for search since the elements in the diagonal covariance matrix $\mathbf{C}$ represent the preferred coordinates of the past successful trials, \ie, larger $c_{ii}$ indicates that searching along the $i$-th coordinate may induce a higher success rate based on the past experience.
According to this, in each iteration we select $k$ $(k\ll m)$ coordinates to generate the random vector $\bm{z}$ with the probability of selecting the $i$-th coordinate being proportional to $c_{ii}$ (in Step 4-5).

\vspace{-1.8ex}
\subsubsection{Dimensionality Reduction}
\label{sec:dimension}
It has been proved that the dimensionality reduction of the search space is useful for acceleration of black-box attacks~\cite{chen2017zoo}.
Based on this, we sample the random vector $\bm{z}$ in a lower dimensional space $\mathbb{R}^{m}$ with $m < n$ (in Step 3).
We then adopt an upscaling operator to project $\bm{z}$ to the original space $\mathbb{R}^n$ (in Step 6).
Note that we do not change the dimension of an input image but only reduce the dimension of the search space. Specifically, we use the bilinear interpolation method as the upscaling operator.

\vspace{-1.8ex}
\subsubsection{Hyper-parameter Adjustment}
\label{sec:alg}

There are also several hyper-parameters in the proposed algorithm, including $\sigma$, $\mu$, $c_c$, and $c_{cov}$. We simply set $c_c=0.01$ and $c_{cov}=0.001$.
$\sigma$ is set as $0.01\cdot\mathcal{D}(\tilde{\bm{x}}^*, \bm{x})$ based on the intuition that $\sigma$ should shrink gradually when the distance from $\bm{x}$ decreases.
$\mu$ is a critical hyper-parameter that needs to be tuned carefully.
If $\mu$ is too large, the search point may probably violate the adversarial criterion and the success rate of updates is low. On the other hand, if $\mu$ is too small, we would make little progress towards minimizing the distance between $\tilde{\bm{x}}^*$ and $\bm{x}$ although the success rate is high.
So we adopt the 1/5th success rule~\cite{rechenberg1978evolutionsstrategien}, which is a traditional method for hyper-parameter control in evolution strategies, to update $\mu$ as $\mu = \mu \cdot \exp(P_{\mathrm{success}} - \nicefrac{1}{5})$,
where $P_{\mathrm{success}}$ is the success rate of several past trials.

\section{Experiments}
In this section, we present the experimental results to demonstrate the effectiveness of the proposed evolutionary attack method. We comprehensively evaluate the robustness of several state-of-the-art face recognition models under the decision-based black-box attack scenario.
We further apply the proposed method to attack a real-world face recognition system to demonstrate its practical applicability.

\begin{table*}[!t]
\footnotesize
\begin{center}
\begin{tabular}{c|c|cccc|cccc|cccc}

\hline
\multicolumn{2}{c|}{Model} & \multicolumn{4}{c|}{SphereFace~\cite{liu2017sphereface}} &  \multicolumn{4}{c|}{CosFace~\cite{wang2018cosface}} & \multicolumn{4}{c}{ArcFace~\cite{deng2018arcface}} \\
\hline
\multicolumn{2}{c|}{Queries} & 1,000 & 5,000 & 10,000 & 100,000 & 1,000 & 5,000 & 10,000 & 100,000 & 1,000 & 5,000 & 10,000 & 100,000 \\
\hline\hline
\multirow{4}{*}{Dodging} & Boundary~\cite{Brendel2018Decision} & 2.4e-2 & 6.5e-3 & 4.7e-4 & 1.4e-5 & 2.0e-2 & 5.1e-3 & 5.4e-4 & 1.2e-5 & 3.1e-2 & 1.7e-2 & 1.6e-3 & 2.3e-5\\
& Optimization~\cite{cheng2018query} & 1.1e-2 & 2.1e-3 & 8.3e-4 & 4.6e-5 & 1.0e-2 & 2.0e-3 & 8.2e-4 & 4.0e-5 & 2.0e-2 & 6.1e-3 & 2.7e-3 & 9.8e-5\\
& NES-LO~\cite{ilyas2018black} & 1.4e-1 & 4.0e-2 & 2.5e-2 & 5.5e-3 & 1.5e-1 & 3.6e-2 & 2.2e-2 & 4.7e-3 & 1.5e-1 & 4.5e-2 & 3.1e-2 & 1.3e-2\\
& Evolutionary & \textbf{1.3e-3} & \textbf{6.6e-5} & \textbf{2.5e-5} & \textbf{9.9e-6} & \textbf{1.2e-3} & \textbf{6.2e-5} & \textbf{2.3e-5} & \textbf{7.5e-6} & \textbf{3.2e-3} & \textbf{1.6e-4} & \textbf{5.4e-5} & \textbf{1.6e-5}\\
\hline
\multirow{4}{*}{Impersonation} & Boundary~\cite{Brendel2018Decision} & 2.4e-2 & 1.1e-2 & 1.7e-3 & 3.6e-5 & 2.5e-2 & 8.9e-3 & 1.3e-3 & 2.3e-5 & 2.5e-2 & 1.3e-2 & 2.5e-3 & 3.8e-5\\
& Optimization~\cite{cheng2018query} & 1.9e-2 & 7.7e-3 & 3.7e-3 & 1.6e-4 & 1.9e-2 & 7.1e-3 & 3.3e-3& 1.1e-4 & 2.0e-2 & 1.1e-2 & 6.0e-3 & 3.5e-4\\
& NES-LO~\cite{ilyas2018black} & 7.9e-2 & 3.8e-2 & 2.8e-2 & 1.0e-2 & 8.8e-2 & 3.7e-2 & 2.7e-2 & 8.8e-3 & 8.8e-2 & 3.4e-2 & 2.3e-2 & 1.1e-2\\
& Evolutionary & \textbf{2.5e-3} & \textbf{1.6e-4} & \textbf{6.3e-5} & \textbf{2.3e-5} & \textbf{2.2e-3} & \textbf{1.3e-4} & \textbf{4.6e-5} & \textbf{1.5e-5} & \textbf{3.7e-3} & \textbf{2.5e-4} & \textbf{8.8e-5} & \textbf{2.6e-5}\\
\hline
\end{tabular}
\end{center}
\vspace{-2ex}
\caption{The results on face identification conducted on the LFW dataset. We report the average distortions (MSEs) of the adversarial images generated by different methods for SphereFace, CosFace, and ArcFace given 1,000, 5,000, 10,000, and 100,000 queries.}
\label{tab:identification}
\vspace{-3.5ex}
\end{table*}

\subsection{Experimental Settings}

\vspace{0.1em}\noindent
{\bf Target models.}
We study three state-of-the-art face recognition models, including SphereFace~\cite{liu2017sphereface}, CosFace~\cite{wang2018cosface} and ArcFace~\cite{deng2018arcface}.
In testing, the feature representation for each image is first extracted by these models. Then, the cosine similarity between feature representations of different images are calculated. Finally, we use the thresholding strategy and nearest neighbor classifier for face verification and identification, respectively.

\vspace{0.1em}\noindent
{\bf Datasets.}
We conduct experiments on the Labeled Face in the Wild (LFW)~\cite{huang2008labeled} and MegaFace~\cite{kemelmacher2016megaface} datasets.
For face verification, in each dataset, we select 500 pairs of face images for dodging attack, in which each pair represent the same identity. And, we select another 500 pairs of face images for impersonation attack, in which the images of each pair are from different identities.
For face identification, in each dataset, we select 500 images of 500 different identities to form a gallery set, and corresponding 500 images of the same identities to form a probe set. We perform dodging and impersonation attacks for images in the probe set. For impersonation attack, the target identity is chosen randomly.
The input image size (\ie, the dimension of the input space $n$) is $112\times112\times3$.
All the selected images can be correctly recognized by the three face recognition models.

\begin{figure}[!t]
  \centering
    \includegraphics[width=1.0\linewidth]{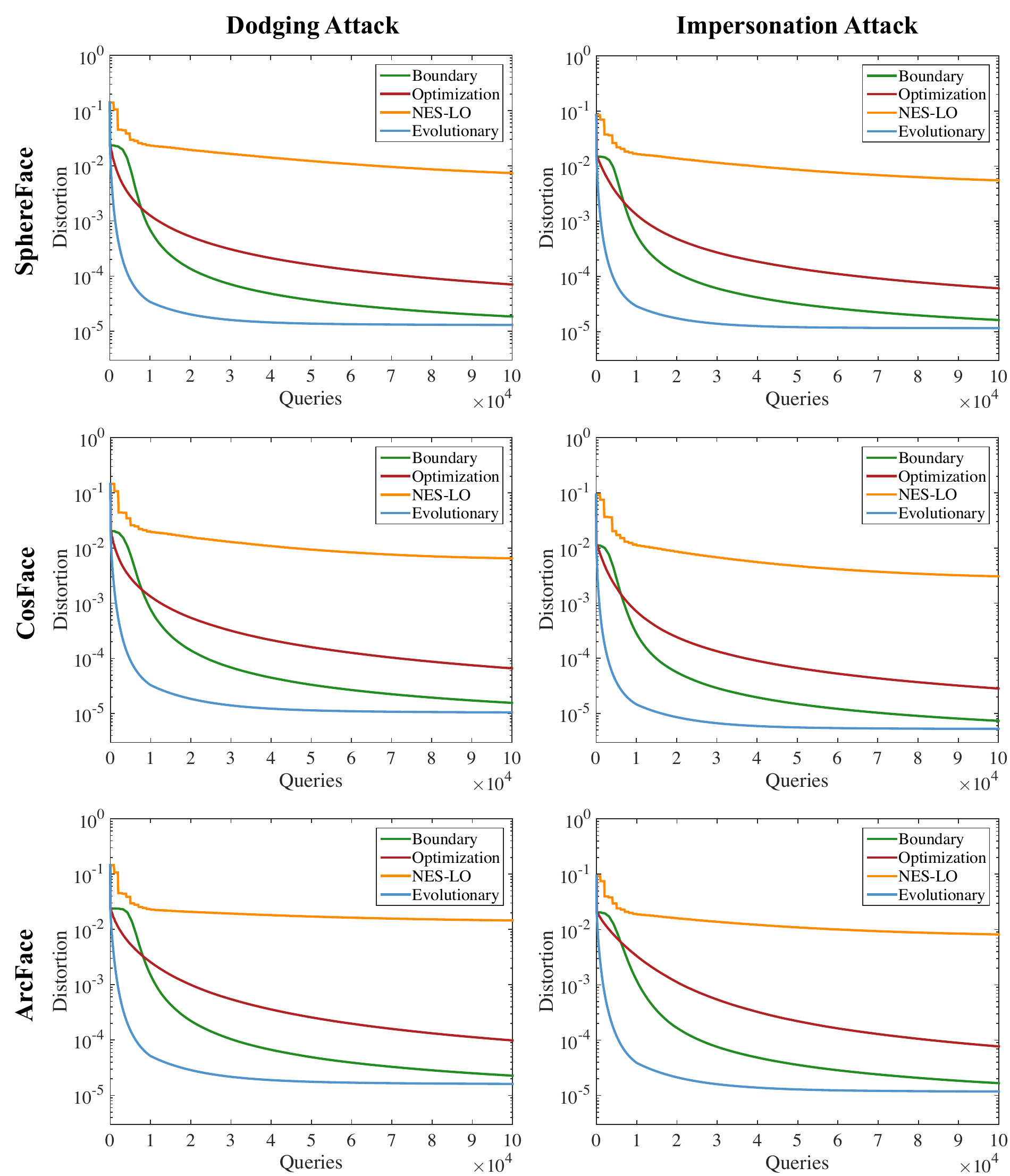}
    \caption{The results on face verification conducted on the LFW dataset. We show the curves of the average distortions (MSEs) of the adversarial images generated by different attack methods for SphereFace, CosFace, and ArcFace over the number of queries.}
    \label{fig:verification}
    \vspace{-2ex}
\end{figure}

\begin{figure}[!t]
  \centering
    \includegraphics[width=1.0\linewidth]{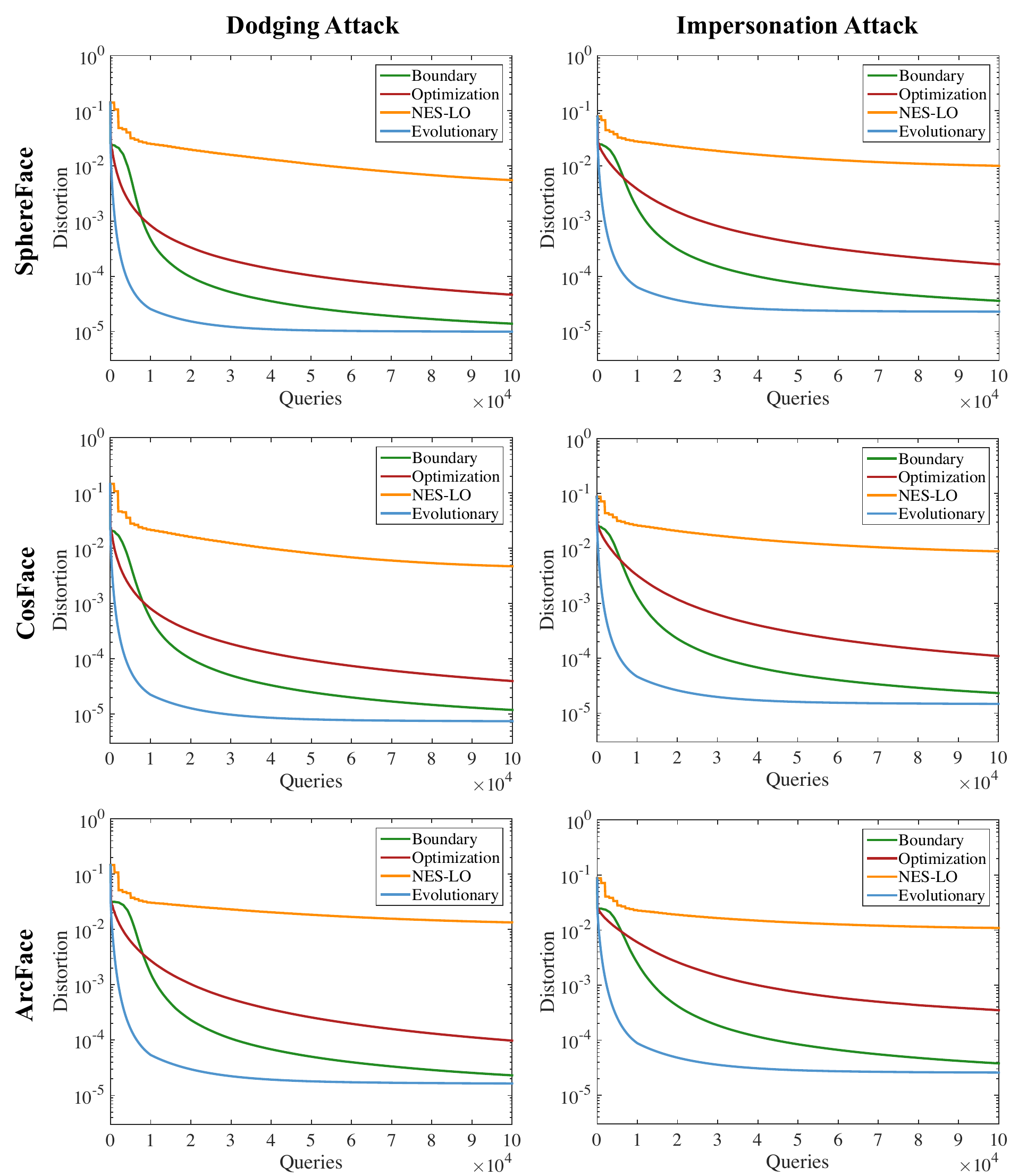}
    \caption{The results on face identification conducted on the LFW dataset. We show the curves of the average distortions (MSEs) of the adversarial images generated by different attack methods for SphereFace, CosFace, and ArcFace over the number of queries.}
    \label{fig:identification}
    \vspace{-2ex}
\end{figure}

\vspace{0.1em}\noindent
{\bf Compared methods.}
We compare the performance of the evolutionary attack method with all existing methods for decision-based black-box attacks, including the boundary attack method~\cite{Brendel2018Decision}, optimization-based method~\cite{cheng2018query} and an extension of NES in the label-only setting (NES-LO)~\cite{ilyas2018black}.

\vspace{0.1em}\noindent
{\bf Evaluation metrics.}
For all methods, the generated adversarial examples are guaranteed to be adversarial. 
So we measure the distortion between the adversarial and original images by mean square error (MSE) to evaluate the performance of different methods\footnote{Images are normalized to $[0,1]$.}. 
We set a maximum number of queries to be 100,000 for each image across all experiments. 
Due to the space limitation, we leave the results on the MegaFace dataset in {\bf Appendix B}.
The results on both datasets are consistent.
Our method is generally applicable beyond face recognition. We further present the results on the ImageNet dataset in {\bf Appendix C}.

\begin{figure*}[t]
  \centering
    \includegraphics[width=0.95\linewidth]{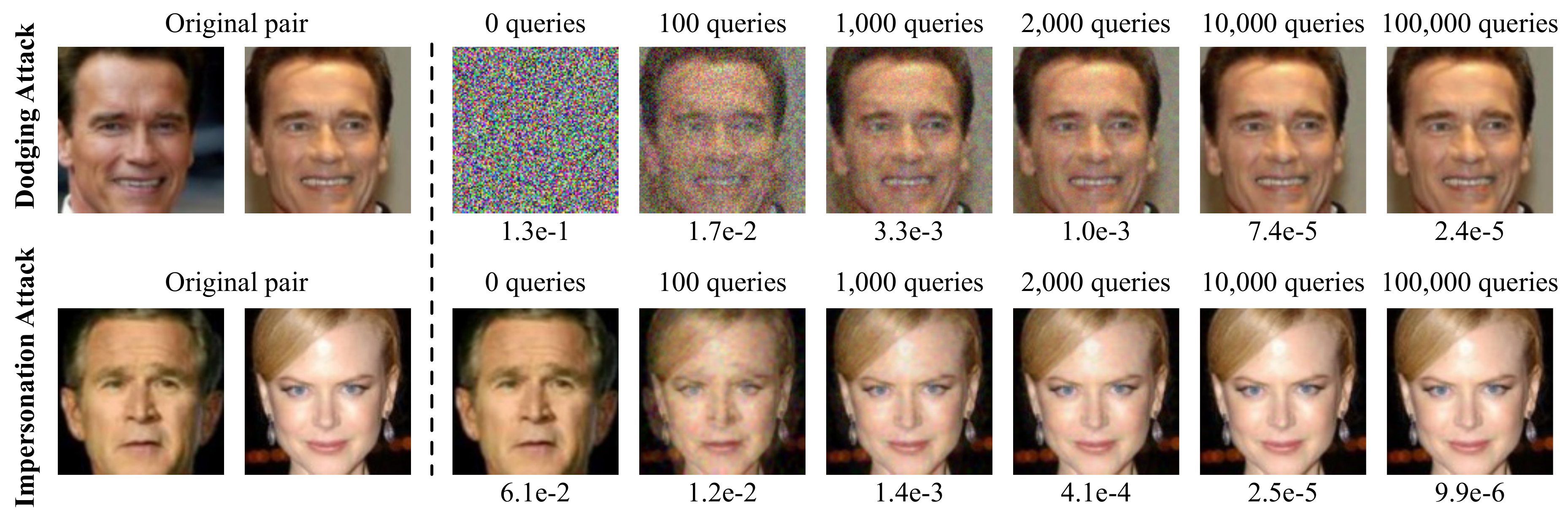}
    \caption{Examples of dodging and impersonation attacks on face verification for the ArcFace~\cite{deng2018arcface} model. The initial adversarial image is a random noise or the target image for each kind of attacks. The distortion between the adversarial image and the original image decreases gradually. We show the total number of queries and the mean square error until each point.}
    \label{fig:result}
    \vspace{-2ex}
\end{figure*}

\subsection{Experimental Results}

We report the results on the LFW dataset in this section.
We perform dodging attack and impersonation attack by Boundary, Optimization, NES-LO, and the proposed Evolutionary method against SphereFace, CosFace, and ArcFace, respectively.
For our method, we set the dimension of the search space as $m=45\times45\times3$, and $k=\nicefrac{m}{20}$ for stochastic coordinate selection.
For other methods, we adopt the default settings.
We calculate the distortions (MSEs) of the adversarial images generated by each method averaged over the selected 500 images. And, the distortion curves over the number of queries for face verification are shown in Fig.~\ref{fig:verification}, while those for face identification in Fig.~\ref{fig:identification}.
Besides, for 1,000, 5,000, 10,000, and 100,000 queries, we report the corresponding distortion values of different methods for face verification in Table~\ref{tab:verification}, while those for face identification in  Table~\ref{tab:identification}. Two visual examples are also presented in Fig.~\ref{fig:result} for dodging and impersonation attacks.
\begin{table}[!t]
\footnotesize
\begin{center}
\begin{tabular}{l|p{11.9ex}<{\centering}|p{11.9ex}<{\centering}|p{11.9ex}<{\centering}}

\hline
& SphereFace & CosFace & ArcFace \\
\hline\hline
wo/ CMA, wo/ SCS &  2.6e-4/1.9e-4 & 2.5e-4/9.2e-5 & 4.2e-4/2.6e-4\\
\hline
w/ CMA, wo/ SCS & 2.4e-4/1.8e-4 & 2.3e-4/8.5e-5 & 3.8e-4/2.5e-4\\
\hline
w/ CMA, w/ SCS ($\mathbf{C}$) & \textbf{1.7e-4/1.3e-4} & \textbf{1.6e-4/6.4e-5} & \textbf{2.6e-4/1.7e-4}\\
\hline
w/ CMA, w/ SCS ($\mathbf{I}_n$) & 2.0e-4/1.5e-4 & 1.9e-4/7.5e-5 & 3.0e-4/2.0e-4\\
\hline
\end{tabular}
\end{center}
\vspace{-2ex}
\caption{Comparisons of the evolutionary method with four settings: without CMA or SCS; with CMA, without SCS; with CMA and SCS where the selection probability is proportional to the elements in $\mathbf{C}$; with CMA and SCS where the selection probability is set equally. We report the average distortions (MSEs) given 10,000 queries for dodging/impersonation attacks on face verification.}
\label{tab:ablation}
\vspace{-4ex}
\end{table}

Above results demonstrate that our method converges much faster and achieves smaller distortions compared with other methods consistently across both tasks (\ie, face verification and identification), both attack settings (\ie, dodging and impersonation), and all face models.
For example, as shown in Table~\ref{tab:verification} and~\ref{tab:identification}, given 5,000 queries our method obtains the distortions which are about 30 times smaller than those generated by the second best method (\ie, Optimization), which validates the effectiveness of the proposed method. 
From Fig.~\ref{fig:result}, it can be seen that 2,000 queries are sufficient to generate visually indistinguishable adversarial examples.
For NES-LO, the hard-label predictions are first used to estimate the predicted probability (\eg, 25 queries) and then it approximates the gradient by NES (\eg, 40 trials). In consequence, this method requires more than 1,000 queries for only one update, which leads to the worst results.

It should be noted that the face recognition models are extremely vulnerable to adversarial examples. These models can be fooled in the black-box manner by adversarial examples with about only $1e^{-5}$ distortions, which are visually imperceptible for humans, as shown in Fig.~\ref{fig:result}.

\subsection{Ablation Study}

\begin{figure}[t]
  \centering
    \includegraphics[width=1.0\linewidth]{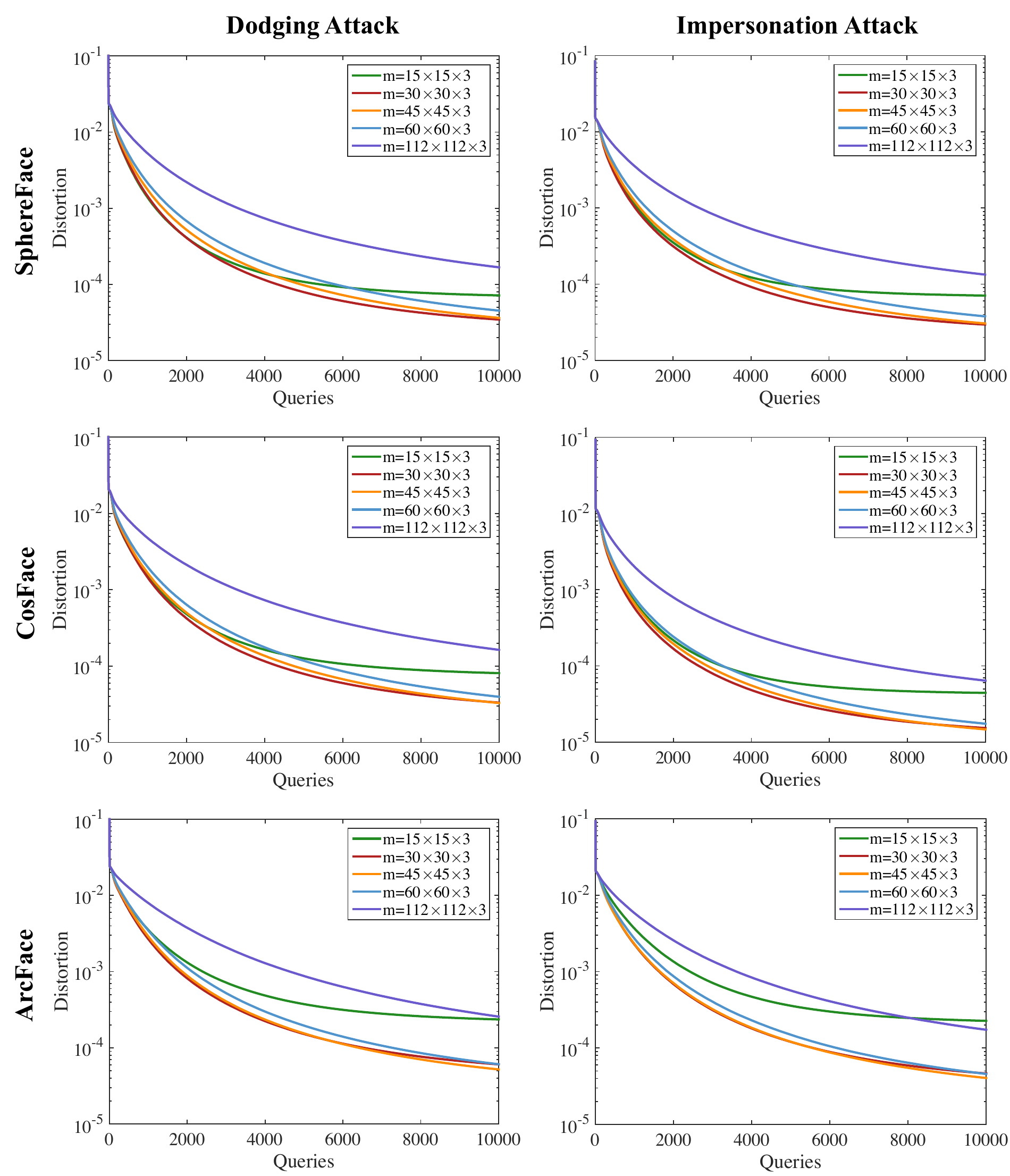}
    \caption{We show the curves of the average distortions (MSEs) of the adversarial images generated by the evolutionary method with different dimensions of the search space over the number of queries. We perform dodging and impersonation attacks against SphereFace, CosFace, and ArcFace on face verification.}
    \label{fig:resize}
    \vspace{-3ex}
\end{figure}

We perform ablation study in this section to validate the effectiveness of each component in the proposed method.
We conduct experiments based on face verification on the LFW dataset.
In particular, we study the effects of covariance matrix adaptation, stochastic coordinate selection and dimensionality reduction respectively.

\textbf{Covariance matrix adaptation (CMA).} To examine the usefulness of CMA, we compare CMA with a baseline method that the covariance matrix is set to $\mathbf{I}_n$ without updating. We do not include stochastic coordinate selection or dimensionality reduction in this part for solely examining the effect of CMA. We show the results of the average distortions given 10,000 queries in the first two rows of Table~\ref{tab:ablation}. CMA improves the results over the baseline method.

\textbf{Stochastic coordinate selection (SCS).} We study two aspects of SCS. The first is whether SCS is useful. The second is whether we should select the coordinates with probability being proportional to the diagonal elements in the covariance matrix $\mathbf{C}$. 
We further perform experiments with SCS, where we compare the performance of SCS with the selection probability of each coordinate being proportional to each diagonal element in $\mathbf{C}$ or $\mathbf{I}_n$ (equal probability for each coordinate).
By comparing the 2-4 rows of Table~\ref{tab:ablation}, it can be seen that SCS is beneficial for obtaining better results and sampling coordinates with probability proportional to $c_{ii}$ is better than sampling with equal probability.

\textbf{Dimensionality reduction.}
We finally study the influence of dimensionality reduction. We set the dimension $m$ of the search space as $15\times15\times3$, $30\times30\times3$, $45\times45\times3$, $60\times60\times3$, and $112\times112\times3$. We perform dodging and impersonation attacks against SphereFace, CosFace, and ArcFace with each $m$, and compare the results in Fig.~\ref{fig:resize}.
It can be seen that the evolutionary method converges faster in a lower dimensional search space. However, if the dimension of the search space is too small (\eg, $15\times15\times3$), the attack results in relatively large distortions. So we choose a medium dimension as $45\times45\times3$ in the above experiments.

\subsection{Attacks on a Real-World Application}

In this section, we apply the evolutionary attack method to the face verification API in Tencent AI Open Platform\footnote{\url{https://ai.qq.com/product/face.shtml\#compare}}.
This face verification API allows users to upload two face images, and outputs a similarity score of them.
We set the threshold to be 90, \ie, if the similarity score is larger than 90, the two images are predicted to be the same identity; and if not, they are predicted to be different identities.

\begin{figure}[t]
  \centering
    \includegraphics[width=0.9\linewidth]{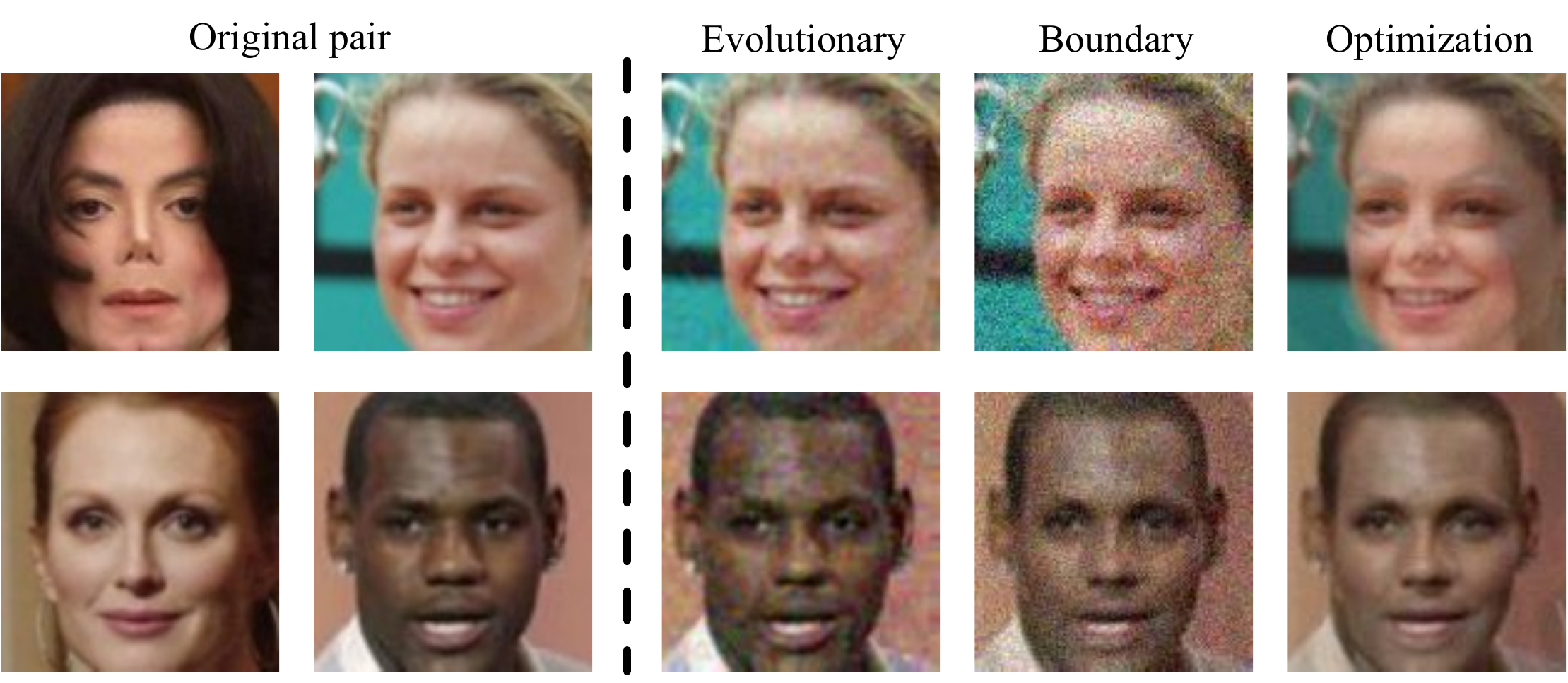}
    \caption{Examples of impersonation attack on the real-world face verification API. We show the original pairs of images as well as the adversarial images generated by each method.}
    \label{fig:api}
    \vspace{-3ex}
\end{figure}

We choose 10 pairs of images from the LFW dataset to perform impersonation attack. The original two face images of each pair are from different identities. We generate a perturbation for one of them and make the API recognize the adversarial image to be the same identity as the other image.
We set the maximum number of queries to be 10,000.
We use the proposed evolutionary method to attack the face verification API and compare the results with Boundary~\cite{Brendel2018Decision} and Optimization~\cite{cheng2018query}.
We do not present the result of NES-LO~\cite{ilyas2018black}, as it fails to generate an adversarial image within 10,000 queries.
We show the average distortion between the adversarial and original images in Table~\ref{tab:api}.
Our method still obtains a smaller distortion than other baseline methods.
We also show two examples in Fig.~\ref{fig:api}.
It can be seen that the adversarial images generated by our method are more visually similar to the original images, while those generated by other methods have large distortions, making them distinguishable from the original images.

\begin{table}[!t]
\begin{center}
\footnotesize
\begin{tabular}{p{20ex}<{\centering}|p{20ex}<{\centering}}

\hline
Attack Method & Distortion (MSE) \\
\hline\hline
Boundary~\cite{Brendel2018Decision} & 1.63e-2 \\
\hline
Optimization~\cite{cheng2018query} & 1.71e-2  \\
\hline
Evolutionary & \textbf{2.54e-3} \\
\hline
\end{tabular}
\end{center}
\vspace{-1ex}
\caption{The results of impersonation attack on the real-world face verification API. We report the average distortions (MSEs) of the selected 10 pairs of images by different attack methods.}
\label{tab:api}
\vspace{-1ex}
\end{table}

\section{Conclusion}

In this paper, we proposed an evolutionary attack method to generate adversarial examples in the decision-based black-box setting.
Our method improves the efficiency over the other methods by modeling the local geometries of the search directions and meanwhile reducing the dimension of the search space.
We applied the proposed method to comprehensively study the robustness of several state-of-the-art face recognition models, and compared against the other methods. The extensive experiments consistently demonstrate the effectiveness of the proposed method.
We showed that the existing face recognition models are extremely vulnerable to adversarial attacks in the black-box manner, which raises security concerns for developing more robust face recognition models.
We finally attacked a real-world face recognition system by the proposed method, demonstrating its practical applicability.

\vspace{-1ex}
\section*{Acknowledgements}
\vspace{-1ex}
Most of this work was done when Yinpeng Dong was an intern at Tencent AI Lab, supported by the Tencent Rhino-Bird Elite Training Program. Yinpeng Dong, Hang Su, and Jun Zhu are supported by the National Key Research and Development Program of China (No. 2017YFA0700904), NSFC Projects (Nos. 61620106010, 61621136008, 61571261), Beijing NSF Project (No. L172037), Tiangong Institute for Intelligent Computing, NVIDIA NVAIL Program, DITD Program JCKY2017204B064, and the projects from Siemens and Intel.

{\small
\bibliographystyle{ieee}
\bibliography{egbib}
}

\clearpage
\setcounter{theorem}{0}
\noindent \begin{center} {\large  \textbf{Appendix}} \end{center}
\appendix
\section{Proof}
\begin{theorem}
Assume that the covariance matrix $\mathbf{C}$ is positive definite. Let $\lambda_{max}$ and $\lambda_{min} (>0)$ be the largest and smallest eigenvalues of $\mathbf{C}$, respectively. Then, we have
\begin{equation*}
P_{\bm{z}\sim\mathcal{N}(\mathbf{0}, \sigma^2\mathbf{C})}\big( \mathcal{L}(\tilde{\bm{x}}^* + \bm{z}) < \mathcal{L}(\tilde{\bm{x}}^*)\big) \leq \frac{4\lambda_{max}\|\tilde{\bm{x}}^*-\bm{x}\|^2}{\sigma^2\lambda_{min}^2n^2}.
\end{equation*}
\end{theorem}

\noindent\textbf{Proof.} Assume that the eigenvalues of the covariance matrix $\mathbf{C}$ are $\lambda_1, \lambda_2, ..., \lambda_n$. Let $\lambda_{max}$ and $\lambda_{min}$ be the largest and smallest eigenvalues, respectively. If $\mathbf{C}$ is positive definite, we have $\lambda_{min} > 0$. Since the covariance matrix $\mathbf{C}$ is a symmetric matrix, we can decompose $\mathbf{C}$ by eigen-decomposition as 
\[
\mathbf{C} = (\mathbf{A}\mathbf{B}) \cdot (\mathbf{A}\mathbf{B})^T,
\]
where $\mathbf{A}$ is an orthogonal matrix and $\mathbf{B}$ is a diagonal matrix whose $i$-th element $b_{ii} = \sqrt{\lambda_i}$.

We assume that $\bm{z}'\sim\mathcal{N}(\mathbf{0}, \mathbf{I})$, and $\bm{z} = \sigma\mathbf{A}\mathbf{B}\bm{z}'$ such that $\bm{z}\sim\mathcal{N}(\mathbf{0}, \sigma^2\mathbf{C})$.
We can then calculate the probability as
\[
\begin{split}
& P_{\bm{z}\sim\mathcal{N}(\mathbf{0}, \sigma^2\mathbf{C})}\big( \mathcal{L}(\tilde{\bm{x}}^* + \bm{z}) < \mathcal{L}(\tilde{\bm{x}}^*)\big) \\
\leq & P_{\bm{z}'\sim\mathcal{N}(\mathbf{0}, \mathbf{I})} \big( \|\tilde{\bm{x}}^* + \sigma\mathbf{A}\mathbf{B}\bm{z}' - \bm{x})\| < \|\tilde{\bm{x}}^* - \bm{x}\|\big) \\
= & P_{\bm{z}'\sim\mathcal{N}(\mathbf{0}, \mathbf{I})} \big( (\tilde{\bm{x}}^* - \bm{x}) \cdot (\sigma\mathbf{A}\mathbf{B}\bm{z}') < -\frac{1}{2}\sigma^2\|\mathbf{A}\mathbf{B}\bm{z}'\|^2 \big) \\
\leq & P_{\bm{z}'\sim\mathcal{N}(\mathbf{0}, \mathbf{I})} \big( (\tilde{\bm{x}}^* - \bm{x}) \cdot (\mathbf{A}\mathbf{B}\bm{z}') < -\frac{1}{2}\sigma \lambda_{min} \|\bm{z}'\|^2 \big).
\end{split}
\]
According to the law of large number~\cite{loeve1977probability}, we have
\[
\|\bm{z}'\|^2 \stackrel{a.s.}{\longrightarrow} n \quad \text{when }  n \rightarrow \infty.
\]
We then calculate the mean and variance of the random variable $\bm{y} = (\tilde{\bm{x}}^* - \bm{x}) \cdot (\mathbf{A}\mathbf{B}\bm{z}') $
\[
\mathrm{E}(\bm{y}) = \mathrm{E}\big[ (\tilde{\bm{x}}^* - \bm{x}) \cdot (\mathbf{A}\mathbf{B}\bm{z}')\big] = 0.
\]
\[
\begin{split}
\mathrm{Var}(\bm{y}) & = \int_{\bm{z}'\sim\mathcal{N}(\mathbf{0}, \mathbf{I})} |(\tilde{\bm{x}}^* - \bm{x}) \cdot (\mathbf{A}\mathbf{B}\bm{z}')|^2 d\bm{z}' \\
& \leq \lambda_{max} \int_{\bm{z}'\sim\mathcal{N}(\mathbf{0}, \mathbf{I})} |(\tilde{\bm{x}}^* - \bm{x}) \cdot \bm{z}'|^2 d\bm{z}' \\
& = \lambda_{max} \|\tilde{\bm{x}}^* - \bm{x}\|^2.
\end{split}
\]
Finally, according to the Chebyshev's inequality~\cite{huber1967behavior}, we have
\[
\begin{split}
& P_{\bm{z}\sim\mathcal{N}(\mathbf{0}, \sigma^2\mathbf{C})}\big( \mathcal{L}(\tilde{\bm{x}}^* + \bm{z}) < \mathcal{L}(\tilde{\bm{x}}^*)\big) \\
\leq & P_{\bm{z}'\sim\mathcal{N}(\mathbf{0}, \mathbf{I})} \big( (\tilde{\bm{x}}^* - \bm{x}) \cdot (\mathbf{A}\mathbf{B}\bm{z}') < -\frac{1}{2}\sigma \lambda_{min} \|\bm{z}'\|^2 \big) \\
 = & P_{\bm{z}'\sim\mathcal{N}(\mathbf{0}, \mathbf{I})} \big(\bm{y} < -\frac{1}{2}\sigma\lambda_{min}n\big) \\
 \leq & P_{\bm{z}'\sim\mathcal{N}(\mathbf{0}, \mathbf{I})} \big(|\bm{y}| > \frac{1}{2}\sigma\lambda_{min}n\big) \\
\leq & \frac{\mathrm{Var} (\bm{y})}{(\frac{1}{2}\sigma\lambda_{min}n)^2} \\
\leq & \frac{4\lambda_{max}\|\tilde{\bm{x}}^* - \bm{x}\|^2}{\sigma^2\lambda_{min}^2n^2}.
\end{split}
\]
\qed

\section{Results on MegeFace}

\begin{figure}[!t]
  \centering
    \includegraphics[width=1.0\linewidth]{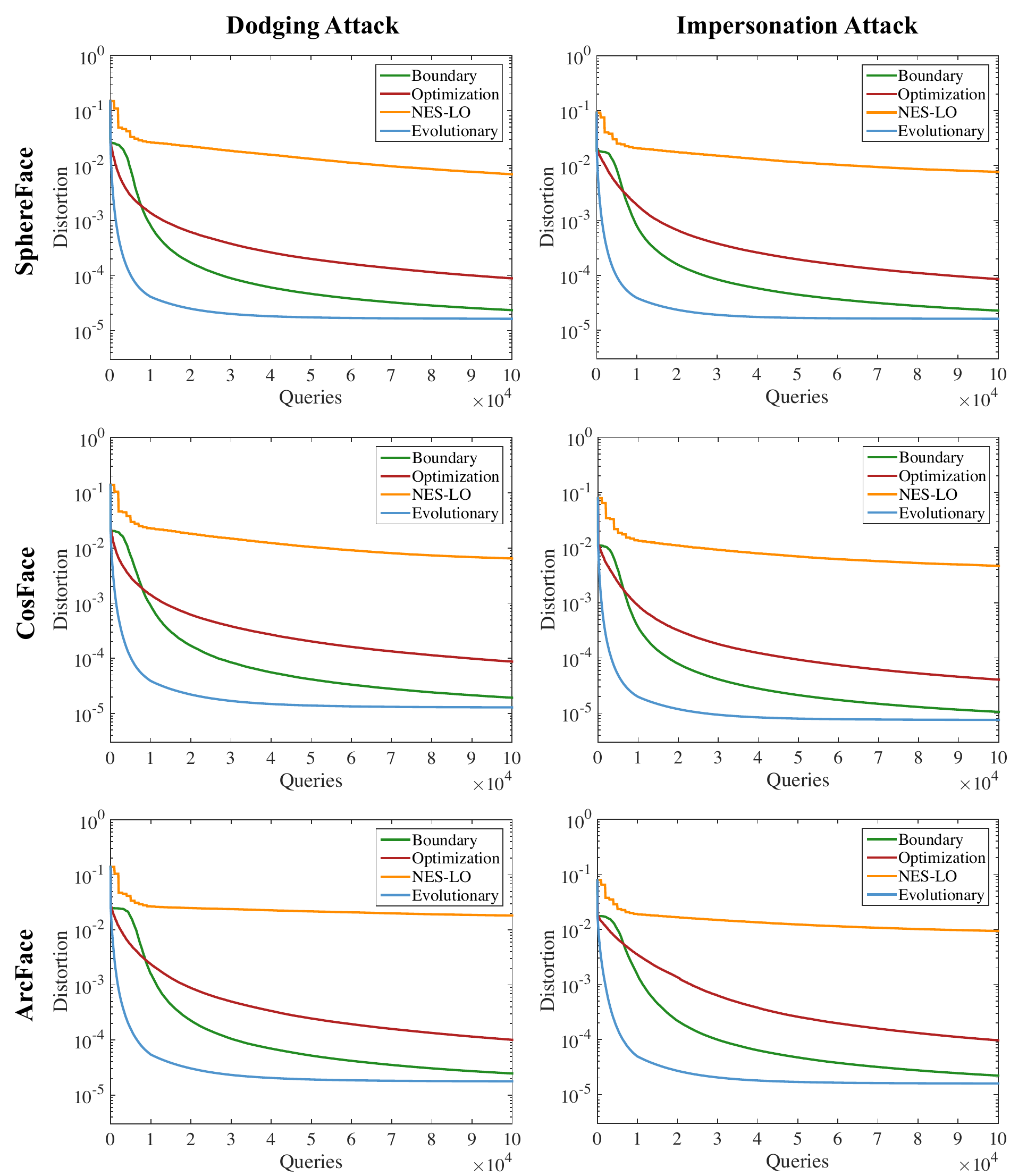}
    \caption{The results on face verification conducted on the MegaFace dataset. We show the curves of the average distortion (MSE) of the adversarial images generated by different attack methods for SphereFace, CosFace, and ArcFace over the number of queries.}
    \label{fig:verification-m}
\end{figure}

\begin{figure}[!t]
  \centering
    \includegraphics[width=1.0\linewidth]{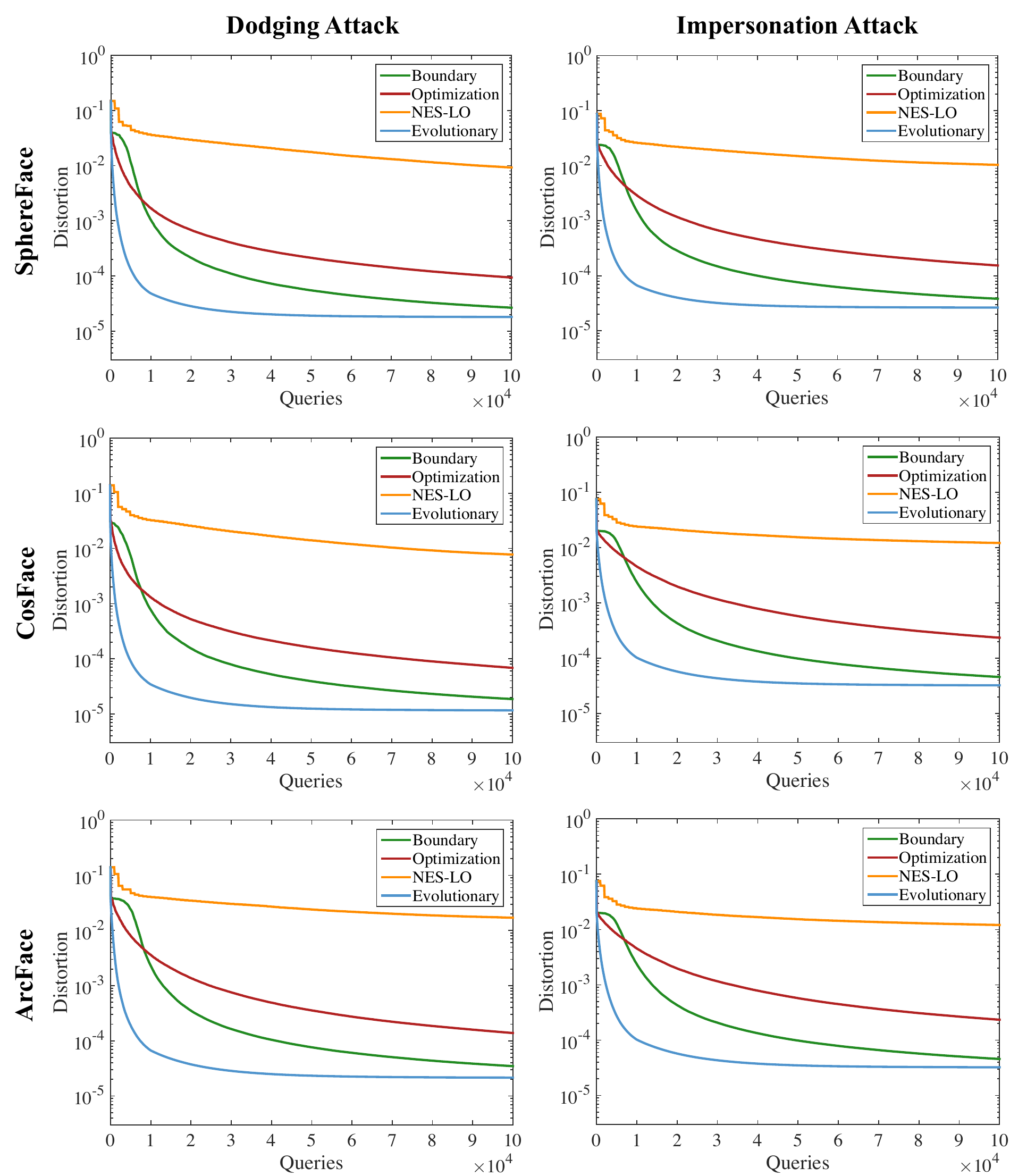}
    \caption{The results on face identification conducted on the MegaFace dataset. We show the curves of the average distortion (MSE) of the adversarial images generated by different attack methods for SphereFace, CosFace, and ArcFace over the number of queries.}
    \label{fig:identification-m}
\end{figure}

We supplement the results on the MegaFace dataset~\cite{kemelmacher2016megaface}. We attack SphereFace~\cite{liu2017sphereface}, CosFace~\cite{wang2018cosface}, and ArcFace~\cite{deng2018arcface} by Boundary~\cite{Brendel2018Decision}, Optimization~\cite{cheng2018query}, NES-LO~\cite{ilyas2018black}, and the proposed Evolutionary for face verification and identification, respectively. We show the distortion curves over the number of queries in Fig.~\ref{fig:verification-m} for face verification, and Fig.~\ref{fig:identification-m} for face identification, respectively.
We also report the distortion values of different methods at 1,000, 5,000, 10,000, and 100,000 queries in Table~\ref{tab:verification-m} for face verification, and Table~\ref{tab:identification-m} for face identification, respectively. 
The proposed method outperforms the other methods in all settings on the MegaFace dataset. The results are consistent with those based on the LFW dataset.

\begin{table*}[!t]
\footnotesize
\begin{center}
\begin{tabular}{c|c|cccc|cccc|cccc}

\hline
\multicolumn{2}{c|}{Model} & \multicolumn{4}{c|}{SphereFace~\cite{liu2017sphereface}} &  \multicolumn{4}{c|}{CosFace~\cite{wang2018cosface}} & \multicolumn{4}{c}{ArcFace~\cite{deng2018arcface}} \\
\hline
\multicolumn{2}{c|}{Queries} & 1,000 & 5,000 & 10,000 & 100,000 & 1,000 & 5,000 & 10,000 & 100,000 & 1,000 & 5,000 & 10,000 & 100,000 \\
\hline\hline
\multirow{4}{*}{Dodging} & Boundary~\cite{Brendel2018Decision} & 2.5e-2 & 8.8e-3 & 8.3e-4 & 2.4e-5 & 2.0e-2 & 7.2e-3 & 9.0e-4 & 1.9e-5 & 2.5e-2 & 1.7e-2 & 1.6e-3 & 2.5e-5\\
& Optimization~\cite{cheng2018query} & 1.3e-2 & 2.9e-3 & 1.4e-3 & 8.9e-5 & 1.1e-2 & 3.0e-3 & 1.4e-3 & 8.7e-5 & 1.7e-2 & 5.3e-3 & 2.4e-3 & 1.0e-4\\
& NES-LO~\cite{ilyas2018black} & 1.5e-1 & 4.2e-2 & 2.7e-2 & 6.9e-3 & 1.4e-1 & 3.8e-2 & 2.3e-2 & 6.5e-3 & 1.4e-1 & 4.2e-2 & 2.7e-2 & 1.8e-2\\
& Evolutionary & \textbf{1.7e-3} & \textbf{1.0e-4} & \textbf{4.1e-5} & \textbf{1.6e-5} & \textbf{1.7e-3} & \textbf{1.0e-4} & \textbf{3.9e-5} & \textbf{1.3e-5} & \textbf{2.6e-3} & \textbf{1.6e-4} & \textbf{5.4e-5} & \textbf{1.8e-5}\\
\hline
\multirow{4}{*}{Impersonation} & Boundary~\cite{Brendel2018Decision} & 1.8e-2 & 8.4e-3 & 7.9e-4 & 2.3e-5 & 1.1e-2 & 3.9e-3 & 3.6e-4 & 1.1e-5 & 1.7e-2 & 9.9e-3 & 1.5e-3 & 2.2e-5\\
& Optimization~\cite{cheng2018query} & 1.4e-2 & 4.6e-3 & 1.9e-3 & 8.5e-5 & 7.7e-3 & 2.3e-3 & 8.9e-4 & 4.0e-5 & 1.4e-2 & 6.7e-3 & 3.5e-3 & 9.6e-5\\
& NES-LO~\cite{ilyas2018black} & 9.2e-2 & 3.0e-2 & 2.1e-2 & 7.7e-3 & 7.9e-2 & 2.2e-2 & 1.4e-2 & 4.7e-3 & 7.9e-2 & 2.9e-2 & 1.9e-2 & 9.3e-3\\
& Evolutionary & \textbf{1.5e-3} & \textbf{9.5e-5} & \textbf{3.9e-5} & \textbf{1.6e-5} & \textbf{8.2e-4} & \textbf{4.9e-5} & \textbf{2.0e-5} & \textbf{7.6e-6} & \textbf{2.7e-3} & \textbf{1.6e-4} & \textbf{4.9e-5} & \textbf{1.6e-5}\\
\hline
\end{tabular}
\end{center}
\caption{The results on face verification conducted on the MegaFace dataset. We report the average distortion (MSE) of the adversarial images generated by different methods for SphereFace, CosFace, and ArcFace given 1,000, 5,000, 10,000, and 100,000 queries.}
\label{tab:verification-m}
\end{table*}

\begin{table*}[!t]
\footnotesize
\begin{center}
\begin{tabular}{c|c|cccc|cccc|cccc}

\hline
\multicolumn{2}{c|}{Model} & \multicolumn{4}{c|}{SphereFace~\cite{liu2017sphereface}} &  \multicolumn{4}{c|}{CosFace~\cite{wang2018cosface}} & \multicolumn{4}{c}{ArcFace~\cite{deng2018arcface}} \\
\hline
\multicolumn{2}{c|}{Queries} & 1,000 & 5,000 & 10,000 & 100,000 & 1,000 & 5,000 & 10,000 & 100,000 & 1,000 & 5,000 & 10,000 & 100,000 \\
\hline\hline
\multirow{4}{*}{Dodging} & Boundary~\cite{Brendel2018Decision} & 3.9e-2 & 1.1e-2 & 1.0e-3 & 2.7e-5 & 2.8e-2 & 7.6e-3 & 7.9e-4 & 1.9e-5 & 3.8e-2 & 2.4e-2 & 2.3e-3 & 3.5e-5\\
& Optimization~\cite{cheng2018query} & 2.0e-2 & 4.2e-3 & 1.7e-3 & 9.4e-5 & 1.4e-2 & 3.0e-3 & 1.3e-3 & 6.9e-5 & 2.6e-2 & 8.0e-3 & 3.6e-3 & 1.4e-4\\
& NES-LO~\cite{ilyas2018black} & 1.5e-1 & 5.3e-2 & 3.7e-2 & 9.3e-3 & 1.4e-1 & 4.7e-2 & 3.3e-2 & 7.7e-3 & 1.4e-1 & 5.5e-2 & 4.1e-2 & 1.7e-2\\
& Evolutionary & \textbf{2.3e-3} & \textbf{1.3e-4} & \textbf{4.8e-5} & \textbf{1.8e-5} & \textbf{1.7e-3} & \textbf{9.3e-5} & \textbf{3.5e-5} & \textbf{1.2e-5} & \textbf{3.6e-3} & \textbf{1.9e-4} & \textbf{6.7e-5} & \textbf{2.2e-5}\\
\hline
\multirow{4}{*}{Impersonation} & Boundary~\cite{Brendel2018Decision} & 2.4e-2 & 1.1e-2 & 1.5e-3 & 3.8e-5 & 2.0e-2 & 7.1e-3 & 1.0e-3 & 2.5e-5 & 2.0e-2 & 1.3e-2 & 2.4e-3 & 4.6e-5\\
& Optimization~\cite{cheng2018query} & 1.7e-2 & 6.1e-3 & 2.9e-3 & 1.5e-4 & 1.4e-2 & 4.7e-3 & 2.1e-3& 1.1e-4 & 1.6e-2 & 8.4e-3 & 4.5e-3 & 2.3e-4\\
& NES-LO~\cite{ilyas2018black} & 8.8e-2 & 3.6e-2 & 2.6e-2 & 1.0e-2 & 7.5e-2 & 3.2e-2 & 2.3e-2 & 8.2e-3 & 7.5e-2 & 3.3e-2 & 2.4e-2 & 1.2e-2\\
& Evolutionary & \textbf{2.4e-3} & \textbf{1.7e-4} & \textbf{6.7e-5} & \textbf{2.6e-5} & \textbf{1.8e-3} & \textbf{1.3e-4} & \textbf{5.0e-5} & \textbf{1.7e-5} & \textbf{3.4e-3} & \textbf{2.7e-4} & \textbf{1.0e-4} & \textbf{3.2e-5}\\
\hline
\end{tabular}
\end{center}
\caption{The results on face identification conducted on the MegaFace dataset. We report the average distortion (MSE) of the adversarial images generated by different methods for SphereFace, CosFace, and ArcFace given 1,000, 5,000, 10,000, and 100,000 queries.}
\vspace{-1ex}
\label{tab:identification-m}
\end{table*}

\begin{figure*}[h]
  \centering
    \includegraphics[width=1.0\linewidth]{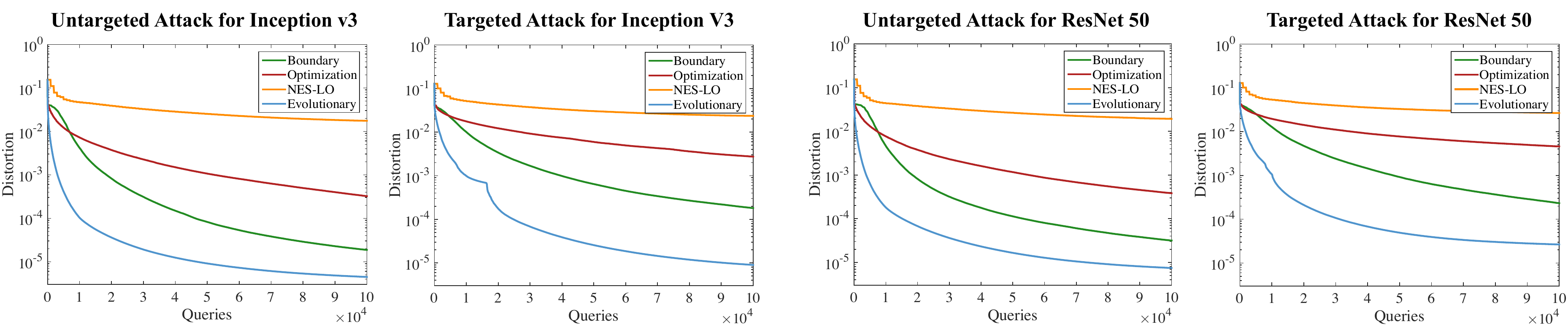}
    \caption{The results of untargeted and targeted attacks on the ImageNet dataset. We show the curves of the average distortion (MSE) of the adversarial images generated by different attack methods for the Inception v3 and ResNet 50 networks over the number of queries.}
    \label{fig:imagenet}
\end{figure*}

\begin{table*}[h]
\begin{center}
\begin{tabular}{c|c|cccc|cccc}

\hline
\multicolumn{2}{c|}{Model} & \multicolumn{4}{c|}{Inception v3~\cite{szegedy2015rethinking}} &  \multicolumn{4}{c}{ResNet 50~\cite{he2015deep}} \\
\hline
\multicolumn{2}{c|}{Queries} & 1,000 & 5,000 & 10,000 & 100,000 & 1,000 & 5,000 & 10,000 & 100,000 \\
\hline\hline
\multirow{4}{*}{Untargeted} & Boundary~\cite{Brendel2018Decision} & 4.0e-2 & 1.8e-2 & 4.2e-3 & 1.9e-5 & 4.1e-2 & 2.1e-2 & 4.6e-3 & 3.2e-5\\
& Optimization~\cite{cheng2018query} & 2.8e-2 & 1.2e-2 & 7.3e-3 & 3.2e-4 & 2.9e-2 & 1.3e-2 & 7.7e-3 & 3.8e-4\\
& NES-LO~\cite{ilyas2018black} & 1.5e-1 & 6.2e-2 & 4.7e-2 & 1.8e-2 & 1.5e-1 & 5.9e-2 & 4.4e-2 & 1.9e-2\\
& Evolutionary & \textbf{5.3e-3} & \textbf{4.2e-4} & \textbf{1.0e-4} & \textbf{4.5e-6} & \textbf{6.6e-3} & \textbf{6.3e-4} & \textbf{1.8e-4} & \textbf{7.4e-6}\\
\hline
\multirow{4}{*}{Targeted} & Boundary~\cite{Brendel2018Decision} & 3.7e-2 & 2.2e-2 & 1.0e-2 & 1.8e-4 & 3.9e-2 & 2.5e-2 & 1.3e-2 & 2.3e-4 \\
& Optimization~\cite{cheng2018query} & 3.4e-2 & 2.3e-2 & 1.8e-2 & 2.7e-3 & 3.6e-2 & 2.5e-2 & 2.0e-2 & 4.6e-3\\
& NES-LO~\cite{ilyas2018black} & 1.3e-1 & 6.6e-2 & 5.2e-2 & 2.4e-2 & 1.3e-1 & 6.7e-2 & 5.4e-2 & 2.7e-2 \\
& Evolutionary & \textbf{1.4e-2} & \textbf{2.7e-3} & \textbf{9.9e-4} & \textbf{9.0e-6} & \textbf{1.6e-2} & \textbf{3.2e-3} & \textbf{1.1e-3} & \textbf{2.7e-5}\\
\hline
\end{tabular}
\end{center}
\caption{The results of untargeted and targeted attacks on the ImageNet dataset. We report the average distortion (MSE) of the adversarial images generated by different methods for the Inception v3 and ResNet 50 networks given 1,000, 5,000, 10,000, and 100,000 queries.}
\label{tab:imagenet}
\end{table*}

\begin{table}[b]
\begin{center}
\begin{tabular}{c|c|c|c|c}
\hline
& & mean & std & max\\
\hline
\multirow{2}{*}{SphereFace~\cite{liu2017sphereface}} &
Dodging & 1.3e-5 & 1.2e-5 & 9.4e-5 \\
& Impersonation & 1.2e-5 & 8.1e-6 & 6.2e-5\\
\hline
\multirow{2}{*}{CosFace~\cite{wang2018cosface}} &
Dodging & 1.1e-5 & 9.4e-6 & 5.2e-5 \\
& Impersonation & 5.3e-6 & 4.3e-6 & 2.4e-5 \\
\hline
\multirow{2}{*}{ArcFace~\cite{deng2018arcface}} &
Dodging & 1.6e-5 & 1.2e-5 & 7.2e-5 \\
& Impersonation & 1.2e-5 & 9.2e-6 & 1.1e-4 \\
\hline
\end{tabular}
\end{center}
\caption{The mean, standard deviation, and maximum of the distortion (MSE) over the 500 pairs of images based on the LFW dataset.}
\label{tab:results}
\end{table}

\section{Results on ImageNet}

It should be noted that the proposed evolutionary attack method is not restricted to attacking face recognition models.
It could be used to perform decision-based black-box attacks for any image classification tasks.
In this section, we conduct additional experiments to demonstrate the effectiveness of the evolutionary attack method in the general object recognition task based on the ImageNet~\cite{russakovsky2015imagenet} dataset.
We use the Inception v3~\cite{szegedy2015rethinking} and ResNet 50~\cite{he2015deep} networks in our experiments. 
We choose 100 images from the ImageNet validation set, which are correctly classified by these two models.
We perform untargeted attack and targeted attack against each model by Boundary, Optimization, NES-LO, and Evolutionary in the decision-based black-box setting.
We show the results in Fig.~\ref{fig:imagenet} and Table~\ref{tab:imagenet}.
The experimental results consistently demonstrate the effectiveness of the proposed method.

\section{Experiments Requested by the Reviewers}
We provide the experimental results requested by the reviewers during the review process.

\subsection{Standard Deviation of the Distortion}

We provide the mean, standard deviation, and maximum of the distortion (MSE) over the 500 pairs of images of LFW in Table~\ref{tab:results}. The results are based on our method for face verification given 100,000 queries. Some adversarial images have larger distortions. But the maximum distortions are smaller than $1.1e^{-4}$, which is almost imperceptible for humans (see the examples in Fig.~\ref{fig:result}).

\subsection{A different Initial Image for Impersonation Attacks}
In impersonation attacks, we use the original target image (enrollment image) as the initialization.
We agree that using a different image of the target identity is more practical than using the enrollment image. However, when we are given an image of the target identity, our method could be always used to find a minimum perturbation, no matter whether the initial image is the enrollment image or a different image. To verify this, we use a different image of the target identity as the initial image to perform impersonation attacks on face verification. The average distortions after 100,000 queries are $1.1e^{-5}$, $4.7e^{-6}$, and $1.1e^{-5}$ for the three models, which are very similar to $1.2e^{-5}$, $5.3e^{-6}$, and $1.2e^{-5}$ shown in Table~\ref{tab:verification}, where the initial image is the enrollment image.

\subsection{Compared with White-box Attacks}
We attack the CosFace model by the white-box attack method PGD~\cite{Madry2017Towards} for face verification. For each pair of face images, we find a minimum perturbation that leads to misclassification by binary search. The average distortions over the 500 pairs are $1.7e^{-5}$ for dodging attack, and $8.0e^{-6}$ for impersonation attack, which are larger than the average distortions given by our method ($1.1e^{-5}$ and $5.3e^{-6}$ shown in Table~\ref{tab:verification}).

\end{document}